\pdfoutput=1

\documentclass[11pt]{article}

\usepackage[preprint]{acl}

\usepackage{times}
\usepackage{latexsym}
\usepackage{times}
\usepackage{latexsym}
\usepackage{multirow}
\usepackage{hyperref} 
\usepackage[T1]{fontenc}
\usepackage{times}
\usepackage{latexsym}
\usepackage{multirow}
\usepackage[T1]{fontenc}

\usepackage[utf8]{inputenc}

\usepackage{microtype}
\usepackage{inconsolata}

\usepackage{graphicx}
\graphicspath{ {figure/} } 

\usepackage{amsmath}
\usepackage{graphicx}
\usepackage{subfigure}
\usepackage{makecell}
\usepackage{asymptote}
\usepackage[multiple]{footmisc}

\usepackage[utf8]{inputenc}

\usepackage{microtype}
\usepackage{inconsolata}

\usepackage{graphicx}
\graphicspath{ {figure/} } 

\usepackage{amsmath}
\usepackage{graphicx}
\usepackage{subfigure}
\usepackage{makecell}
\usepackage{asymptote}

\usepackage[T1]{fontenc}

\usepackage[utf8]{inputenc}

\usepackage{microtype}

\usepackage{inconsolata}
\usepackage{booktabs}
\usepackage{graphicx}
\graphicspath{{images/}}

\title{F$^{2}$DR: A Fine-Grained Full-Pipeline Reward Framework for DeepSearch Workflows}

\author{
\textbf{Bojian Xiong}\textsuperscript{\rm{1},*},
\textbf{Wentao Ding}\textsuperscript{\rm{2},*},
\textbf{Yujing Lu}\textsuperscript{\rm{2}},
\textbf{Shaowei Zhang}\textsuperscript{\rm{1}},
\textbf{Ling Shi}\textsuperscript{\rm{1}}, \\
\textbf{Jing Liao}\textsuperscript{\rm{2}},
\textbf{Yan Wang}\textsuperscript{\rm{2}},
\textbf{Yueyang Zhang}\textsuperscript{\rm{2}},
\textbf{Long Xia}\textsuperscript{\rm{2}},
\textbf{Zhiyuan Sun}\textsuperscript{\rm{2}}, \\
\textbf{Daiting Shi}\textsuperscript{\rm{2}},
\textbf{Jingzhou He}\textsuperscript{\rm{2}},
\textbf{Yuqi Ren}\textsuperscript{\rm{1},\dag},
\textbf{Deyi Xiong}\textsuperscript{\rm{1},\dag} \\
\textsuperscript{1}TJUNLP Lab, Tianjin University, Tianjin, China \\
\textsuperscript{2}Baidu Inc., Beijing, China \\
\texttt{\{xbj1355, dyxiong\}@tju.edu.cn}
}

\begin{document}

\maketitle

\begingroup
\renewcommand{\thefootnote}{\fnsymbol{footnote}}
\footnotetext[1]{These authors contributed equally to this work.}
\footnotetext[2]{Corresponding authors.}
\endgroup

\begin{abstract}

With the widespread industrial deployment of Large Language Models (LLMs), DeepSearch has emerged as the dominant paradigm for resolving complex user queries. It typically operates through an iterative closed-loop workflow consisting of planning and reflection, information retrieval, and answer generation. However, existing reward models (RMs) and evaluation benchmarks are primarily designed for static single-turn tasks, failing to capture the full-pipeline complexity of DeepSearch workflows. To address this limitation, we propose \textbf{F$^{2}$DR}, a fine-grained full-pipeline DeepSearch reward framework. F$^{2}$DR evaluates DeepSearch workflows across three dimensions: Content, Trajectory, and Answer, enabling comprehensive process-level assessment. We further construct \textbf{DeepSearch RM-Bench}, a dedicated benchmark for evaluating RMs in DeepSearch scenarios. Extensive experiments demonstrate that F$^{2}$DR achieves significantly higher evaluation consistency than self-evaluation-based baselines, while DeepSearch RM-Bench exhibits strong discriminative capability across existing open-source RMs. We will publicly release the complete DeepSearch RM-Bench dataset soon.

\end{abstract}

\section{Introduction}
With the industrial-scale deployment of large language models (LLMs), user demands for information acquisition have shifted from simple factoid retrieval to complex knowledge-intensive services. In this context, DeepSearch \cite{DBLP:journals/corr/abs-2510-24701, DBLP:journals/corr/abs-2508-12800, DBLP:journals/corr/abs-2511-07327, DBLP:journals/corr/abs-2511-11793, zhou2026deepresearchpretrainingpredictive} has emerged as a dominant paradigm for autonomous information exploration through iterative cycles of planning-reflection, information retrieval, and answer generation. Unlike single-turn question answering systems \cite{DBLP:journals/corr/abs-2509-07968}, DeepSearch operates as a multi-stage workflow involving long-horizon reasoning, tool invocation, and multi-source evidence integration, thereby introducing fundamental challenges for evaluation. Accurately assessing such workflows requires evaluation methods capable of capturing multiple dimensions of system behavior. In industrial practice, generation quality is typically assessed through automated feedback provided by reward models (RMs). However, existing RMs are predominantly designed for static or single-step scenarios, falling short of the process-level evaluative requirements of DeepSearch scenarios.

Existing RMs mainly fall into two paradigms, both of which are inadequate for DeepSearch. Discriminative RMs \cite{DBLP:journals/corr/abs-2410-18451, DBLP:journals/corr/abs-2507-01352, DBLP:journals/corr/abs-2403-17297, DBLP:journals/corr/abs-2406-12845} provide only scalar scores without explicit reasoning, failing to interpret complex multi-step planning. Generative RMs \cite{DBLP:journals/corr/abs-2504-02495, DBLP:journals/corr/abs-2505-02387, DBLP:journals/corr/abs-2506-03637, DBLP:conf/acl/QinLLWWZXSLS26, DBLP:conf/acl/LiangLWWZXSS26}, despite their interpretability, are prone to hallucinations and poor adaptability, failing to meet strict factual accuracy requirements. Beyond the limitations of RMs, existing RM benchmarks are also poorly aligned with DeepSearch workflows \cite{DBLP:journals/corr/abs-2403-13787, DBLP:journals/corr/abs-2506-01937, DBLP:journals/corr/abs-2410-16184}. They primarily focus on well-bounded static scenarios such as general QA and elementary math reasoning, which differ from the ambiguous, multi-dimensional user intents in real DeepSearch applications.

To address these limitations, we propose \textbf{F$^{2}$DR}, a fine-grained full-pipeline reward framework for evaluating DeepSearch workflows. F$^{2}$DR performs a comprehensive assessment across three dimensions: \textbf{Content} dimension assesses the semantic coverage and factual grounding of retrieved information against core knowledge points, utilizing fine-grained checklists to verify factual and data consistency; \textbf{Trajectory} dimension evaluates multi-turn reasoning and task decomposition through six rigorous sub-dimensions, including intent understanding, search atomicity, planning consistency, gap resolution, planning comprehensiveness, and iterative innovation, while penalizing redundant reasoning and ensuring alignment between planning and execution; \textbf{Answer} dimension combines subjective criteria, such as intent matching, logical coherence, and conciseness, with objective metrics measuring core information coverage, thereby ensuring both factual completeness and logical rigor in final responses. With these three dimensions, F$^{2}$DR delivers holistic and process-level evaluation for DeepSearch workflows.


Built upon F$^{2}$DR, we further construct \textbf{DeepSearch RM-Bench}, a dedicated benchmark for reward model evaluation in full-pipeline DeepSearch scenarios. Specifically, during data construction, we adopt a ``planning-reflection -- information acquisition -- answer generation'' workflow. In the planning-reflection phase, the system uses a Directed Acyclic Graph (DAG) to decompose queries into atomic sub-tasks and dynamically adjusts search strategies via metacognitive reflection on evidence sufficiency. During information-acquisition phase, it executes the DAG by integrating search engines and Model Context Protocol (MCP) tools to gather multi-source data, ensuring retrieval breadth and depth. Finally, in the answer generation phase, it aggregates cross-turn evidence for fact verification and conflict resolution, optimizing the reasoning chain to produce a high-quality structured response. To maintain an appropriate level of evaluation difficulty, we filtered out preference pairs with excessively high or low score variances and conduct secondary human verification to ensure sufficient discriminability across both planning trajectories and final responses for identical queries. 
Extensive experiments demonstrate that F$^{2}$DR provides substantially more consistent and reliable evaluations than self-evaluation-based baselines. Furthermore, mainstream open-source RMs exhibit significant performance bottlenecks on DeepSearch RM-Bench, indicating the substantial difficulty and discriminative capability of our benchmark.

In summary, our contributions are as follows: 

\begin{itemize}
    \item 
    We propose F$^{2}$DR, a process-level reward framework tailored for complex real-world DeepSearch workflows, covering three complementary dimensions: Content, Trajectory, and Answer.
    \item 
    We develop DeepSearch RM-Bench, a high-quality benchmark dedicated to assessing RMs in full-pipeline DeepSearch scenarios.
    \item
    Through extensive experiments, we demonstrate that F$^{2}$DR achieves substantially higher evaluation consistency, while current state-of-the-art reward models still exhibit considerable limitations on DeepSearch RM-Bench.
\end{itemize}

\begin{figure*}[!t] 
  \centering  
  \includegraphics[width=1\textwidth]{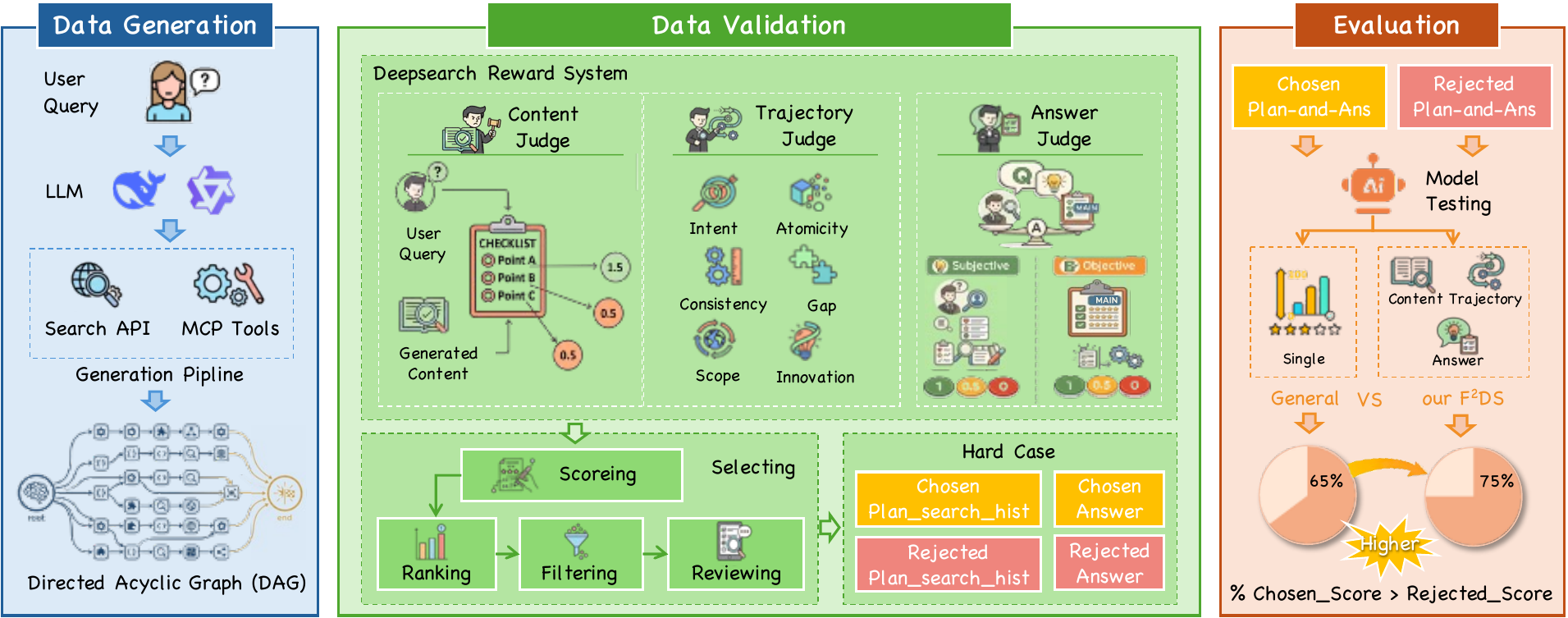}  
\caption{Full-pipeline overview of DeepSearch RM-Bench based on F$^{2}$DR, consisting of three phases: (1) Data Collection: generating full-pipeline DeepSearch trajectories by LLMs equipped with search tools; (2) DeepSearch RM-Bench Construction: conducting  fine-grained full-pipeline evaluation through the F$^{2}$DR framework, followed by filtering and human verification to construct DeepSearch RM-Bench; (3) Evaluation: benchmarking RMs performance and validating F$^{2}$DR superiority.}
\label{fig:fig_1} 
\end{figure*}

\section{Related Work}

\noindent \textbf{DeepSearch.} DeepSearch integrates the reasoning capabilities of LLMs with search engines and tool invocations to construct an iterative workflow, comprising ``planning-reflection, information retrieval, and answer generation'', to address complex tasks in real-world scenarios. Current optimization strategies predominantly focus on workflow-based prompt engineering \citep{DBLP:journals/corr/abs-2501-05366}, supervised fine-tuning \citep{DBLP:journals/corr/abs-2505-20285}, and end-to-end reinforcement learning for autonomous agents \citep{DBLP:journals/corr/abs-2508-13167, DBLP:journals/corr/abs-2507-19849, DBLP:journals/corr/abs-2508-20368, DBLP:journals/corr/abs-2509-06283, DBLP:journals/corr/abs-2601-03743}. However, existing research lacks a specialized evaluation workflow for DeepSearch; current assessments largely rely on general LLM-as-a-judge approaches, which are often ill-equipped to conduct granular assessments of the complex issues within DeepSearch scenarios.

\noindent \textbf{RMs and RM Benchmarks.} Existing RMs can be categorized into two paradigms. Discriminative RMs \citep{DBLP:journals/corr/abs-2406-08673, DBLP:journals/corr/abs-2404-02078}, typically built upon the Bradley-Terry (BT) framework, represent relative preferences by outputting scalar scores for response pairs. However, such methods provide limited interpretability and lack explicit reasoning processes. While generative RMs \citep{DBLP:journals/corr/abs-2505-14674, DBLP:journals/corr/abs-2509-02492} can formulate scoring principles, they often suffer from poor adaptability to specific contexts and are prone to factual hallucinations. In terms of benchmarking, existing RM benchmarks \citep{DBLP:conf/iclr/ZhouZWXDBSXFMZG25, DBLP:journals/corr/abs-2410-12784} focus on static scenarios, such as mathematics and general question-answering. They fail to adequately cover the complex decision-making trajectories and dynamic interactions inherent in real-world DeepSearch workflows.

\section{Methodology}
We first filter and construct high-quality DeepSearch query datasets. Then we propose F$^{2}$DR, a fine-grained full-pipeline framework for evaluating DeepSearch. Based on real-world industrial search data, we further construct DeepSearch RM-Bench by annotating each stage of workflow with F$^{2}$DR. Figure~\ref{fig:fig_1} illustrates the construction process of DeepSearch RM-Bench with F$^{2}$DR.

\subsection{DeepSearch Data Collection}


Our DeepSearch query is drawn from a large-scale enterprise search platform. To select high-quality queries from massive user logs, we employ a dual-stage filtering mechanism. First, we extract raw user queries explicitly invoking "deep thinking", then utilize DeepSeek-V3 \citep{DBLP:journals/corr/abs-2412-19437} to filter out trivial or single-hop queries, retaining only those that require complex, multi-hop reasoning.


For these retained complex queries, we replicate the full industrial DeepSearch execution pipeline to generate diverse reasoning trajectories and final responses. DeepSearch addresses such knowledge-intensive tasks through an iterative closed-loop workflow of planning-reflection, information retrieval, and answer generation. To handle these queries effectively, we adopt a Directed Acyclic Graph (DAG)-based planning-reflection paradigm, departing from the linear chain-based approaches used in prior work such as Search-R1 \citep{DBLP:journals/corr/abs-2503-09516, DBLP:journals/corr/abs-2503-19470, DBLP:journals/corr/abs-2503-05592, DBLP:conf/emnlp/LengLLZXZ0W0X25, DBLP:conf/emnlp/XiongLLZZLL0ZZ025}. Unlike linear paradigms that rely on strictly sequential iterations, the DAG structure models reasoning as a graph: each node represents an atomic sub-goal involving search APIs or MCP tool calls, and directed edges explicitly encode dependencies between sub-tasks. Retrieval results from completed sub-goals directly inform subsequent planning steps, enabling continuous re-planning and real-time DAG updates. This globally coordinated task decomposition and dynamic dependency tracking enable our workflow to construct rigorous execution trajectories for multi-hop problems, effectively identifying and bridging information gaps across all planning stages.

In the final response stage, we synthesize the information retrieved across multiple turns into a response with logical coherence and factual completeness. To construct high-quality preference data, we employ DeepSeek-V3 and Qwen-Next-80B-A3B as LLMs within the DeepSearch workflow. Their distinct reasoning trajectories and retrieval behaviors yield diverse outputs for identical queries, forming reliable preference pairs for reward model evaluation. A detailed case study of the workflow is provided in Appendix \ref {appendixA}.

\subsection{F$^{2}$DR}
To facilitate automated evaluation and defect identification for DeepSearch workflows, we design a fine-grained full-pipeline reward framework comprising Content, Trajectory, and Answer dimensions. These dimensions enable comprehensive assessment across the entire DeepSearch workflow, including planning-reflection, information retrieval, and answer generation. Detailed prompts are provided in Appendix \ref{appendixB}.

\paragraph{Content.} This dimension evaluates the semantic coverage and factual grounding of retrieved information against knowledge points. Specifically, we aggregate responses from frontier models (e.g., Gemini\footnote{\url{https://gemini.google.com/app}}, Doubao\footnote{\url{https://www.doubao.com/chat/}}, and Qwen\footnote{\url{https://qwen.ai/home/}}) operating in web-search environments to obtain high-confidence, multi-source reference answers. Then, DeepSeek-R1 \citep{DBLP:journals/corr/abs-2501-12948} is utilized to summarize and decompose these results into atomic knowledge points defined as \textit{(Entity, Attribute/State)} pairs. These units are further categorized into core and auxiliary knowledge points, forming a checklist for fine-grained verification. Evaluation is subsequently performed by measuring both the semantic coverage and factual consistency of retrieved evidence against this knowledge inventory.


\paragraph{Trajectory.} This dimension performs hierarchical logical auditing of the dynamic reasoning process during the planning-reflection phase, ensuring the rigor of the entire reasoning pipeline at both the task and trajectory levels. We conduct in-depth analysis on execution logs and identify common flaws in DeepSearch reasoning chains, including mixed search keywords, disjointed planning and DAG execution, and repetitive operations. We refine these defects and categorize them into task-level and trajectory-level evaluation dimensions. At the task level, we evaluate the rationality of individual subtasks through three core metrics: intent understanding correct, which verifies the model's accuracy in recognizing both explicit user needs and implicit constraints; search task atomicity, which verifies the compliance of each search task as a semantically independent minimal unit to prevent cross-topic mixed searches; and reasoning-planning consistency, which verifies the alignment between information gaps identified during reasoning and corresponding concrete search tasks. At the trajectory level, guided by the core principle of continuous information entropy reduction, meaning each search round effectively reduces uncertainty, we assess the overall quality of the entire multi-turn reasoning chain through three core metrics: gap resolution degree, which quantifies the proportion of identified information gaps that have been filled; planning comprehensiveness, which verifies the completeness of the plan in covering all information dimensions required to answer the query; and reasoning iterative innovation, which evaluates the ability of each search round to bring new incremental information, thereby avoiding meaningless repetitive exploration. This hierarchical evaluation scheme enables precise localization of specific defects within the planning-reflection pipeline.

\paragraph{Answer.} This dimension integrates subjective qualitative assessment with checklist-based objective quantitative metrics. Subjectively, we evaluate intent alignment, logical coherence, and response conciseness. Objectively, we quantify answer completeness by measuring the coverage rate of core knowledge points from the pre-defined checklist. This hybrid approach enables a robust evaluation of the final response.

Based on DeepSearch ablation experiments, the following weights yield optimal performance (Table \ref{tab:dimension_weight}). The final reward score is:
\begin{equation*}
\begin{split}
\text{Score}_{\text{final}} &= 0.75 \cdot \text{Score}_{\text{content}} + \text{Score}_{\text{trajectory}} \\
&\quad + 1.25 \cdot \text{Score}_{\text{answer}}
\end{split}
\end{equation*}

\begin{table}[t]  
\centering
\footnotesize 
\setlength{\tabcolsep}{2.5pt}  
\renewcommand{\arraystretch}{0.8}  
\begin{tabular}{@{}llc@{}}  
\toprule
\textbf{Dimension} & \textbf{Sub-dimension} & \textbf{Consistency} \\
\midrule
Content 
& Checklist Coverage Rate & 0.9303 \\
\midrule
\multirow{6}{*}{Trajectory}
& Intent Understanding Correctness & 0.9662 \\
& Search Task Atomicity & 0.9114 \\
& Reasoning-Planning Consistency & 0.9485 \\
& Gap Resolution Degree & 0.9627 \\
& Planning Comprehensiveness & 0.9301 \\
& Reasoning Iteration Innovation & 0.9573 \\
\midrule
\multirow{3}{*}{Answer}
& Demand Understanding & 0.9542 \\
& Content Quality & 0.9402 \\
& Answer Coverage Rate & 0.9293 \\
\bottomrule
\end{tabular}
\caption{Human consistency results across different evaluation dimensions.}
\label{tab:human_consistency}
\end{table}

\subsection{DeepSearch RM-Bench Construction}

\paragraph{Data Filtering.} We propose a preference data filtering strategy that integrates multi-model consensus and collaborative expert verification to construct DeepSearch RM-Bench with high discriminability and appropriate difficulty. It is important to distinguish two levels of filtering. At the query level, we remove only trivial and single-hop queries and retain every question that requires multi-hop reasoning, so no hard question is dropped at this stage. At the preference-pair level, advanced reasoning models, including DeepSeek-R1, Doubao-Seed-1.6-Thinking, and Qwen-3-235B-A22B-Thinking, independently score each sample multiple times, with preference polarity determined via majority voting to mitigate model family bias. To balance discriminability and task difficulty, we then filter out pairs with excessively large or small score gaps: a too-large gap indicates that one response is clearly better and hence too easy, while a too-small gap is what we refer to as ambiguous—note that this term describes preference pairs, not the queries themselves. For these small-gap ambiguous pairs, we further conducted a round of manual re-review and found that they are difficult to distinguish not only for the LLMs but also for our human annotators, who likewise could not clearly tell which of the two responses was better; in other words, they are genuine ties, and we remove them because the preference direction itself is unreliable and retaining them would only introduce noise. The score gaps of the retained pairs are generally not large (with an average gap of about 0.2774), which indicates that we intentionally keep a substantial number of closely matched, challenging pairs. We prioritize samples with balanced performance across Content, Trajectory, and Answer dimensions, ensuring logical consistency across planning, reasoning, and final answers.

\begin{figure}[!t] 
  \centering
  \includegraphics[width=1\linewidth]{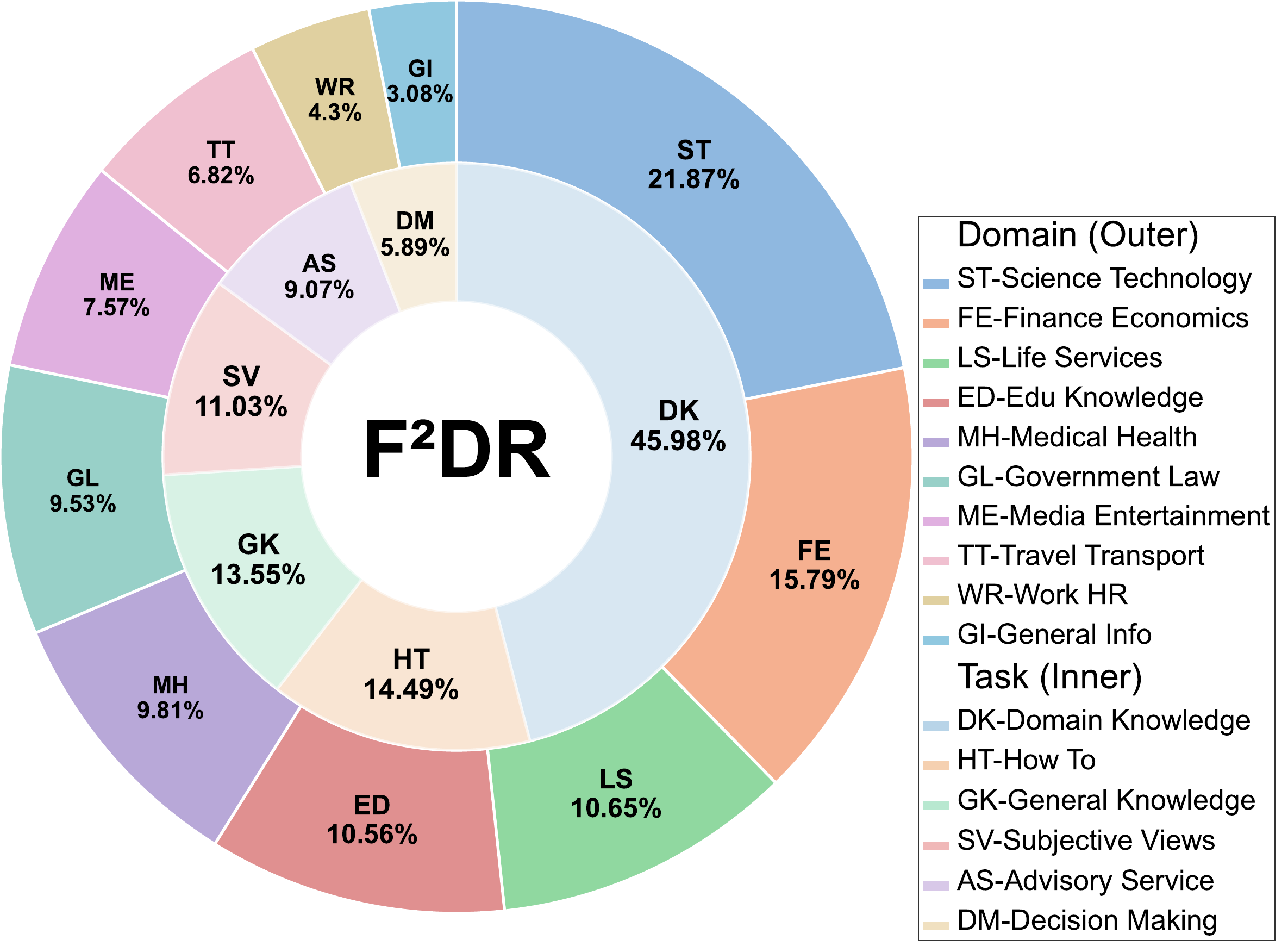}  
  \caption{Task(Inner) and Domain(Outer) Distribution of the DeepSearch RM-bench.}  
  \label{fig:fig_2}
\end{figure}

\paragraph{Manually Verification.} Systematically trained internal annotators manually verify the automatically generated checklists and model scoring rationales according to F$^{2}$DR guidelines, while filtering out invalid cases caused by retrieval failures from search engines. We evaluate inter-annotator agreement using Fleiss' Kappa coefficient. The coefficients across all dimensions exceed 0.9, indicating almost perfect consistency among annotators. Such high consistency is attributed to the annotators' domain expertise and rigorous training procedures, alongside comprehensive annotation guidelines that explicitly define evaluation criteria and edge cases. Table \ref{tab:human_consistency} summarizes the agreement statistics across the Content, Trajectory, and Answer dimensions.


\paragraph{Quality Evaluation.} 
Through the multi-stage filtering pipeline, we obtain a candidate pool of preference pairs. Since the F$^{2}$DR framework is used to filter this data, relying on it again for evaluation would create a self-confirmation loop—essentially allowing the system to act as both player and referee, inherently assigning artificially high scores to its own selections. To break this loop and objectively validate the dataset quality, we introduce an independent, human-led industrial AI search evaluation system as a third-party benchmark. This evaluation system employs a five-dimensional scoring scheme: intent comprehension, factual accuracy, information utility, professionalism, and contextual relevance. Five domain experts in deepsearch conducted a blind review of the candidate pairs. Detailed information on annotators can be found in Appendix \ref{appendixG}. Specifically designed to evaluate the validity of preference pairs, the assessment follows the standard Side-by-Side protocol using Good-Same-Bad (GSB) logic: a preference pair is labeled \textit{Good} if the chosen response outperforms the rejected one, \textit{Same} if their quality is indistinguishable, and \textit{Bad} if the chosen response is inferior. The detailed design and full results of this GSB evaluation are provided in Appendix \ref{appendixE}.

\begin{table}[!t] 
\centering
\renewcommand{\arraystretch}{0.85}
\footnotesize 
\setlength{\tabcolsep}{3pt} 
\begin{tabular}{lcccc}
\toprule
\textbf{Path Type} & \makecell{\textbf{Planning} \\ \textbf{Rounds}} & \makecell{\textbf{Search} \\ \textbf{Counts}} & \makecell{\textbf{Tokens} \\ \textbf{(Plan)}} & \makecell{\textbf{Tokens} \\ \textbf{(Ans)}} \\
\midrule
Chosen Path & 3.63 & 61.56 & 22890.56 & 1136.83 \\
Rejected Path & 3.45 & 51.59 & 18838.64 & 951.60  \\
\bottomrule
\end{tabular}
\caption{Comparison of execution statistics between Chosen and Rejected paths.}
\label{tab:compare_chosen_rejected}
\end{table}

\begin{table*}[!t]
\centering
\renewcommand{\arraystretch}{0.85}
\setlength{\abovetopsep}{0pt}       
\setlength{\belowbottomsep}{0pt}    
\setlength{\aboverulesep}{1pt}       
\setlength{\belowrulesep}{1pt}       
\vspace{-3pt}
\resizebox{\textwidth}{!}{%
\begin{tabular}{l c c | l c}
\toprule
\textbf{Model} & \textbf{Self-evaluation} & \textbf{F$^{2}$DR} & \textbf{Model} & \textbf{preference accuracy} \\
\midrule
GPT-5 & 58.97 & \textbf{69.25} & Skywork-Reward-V2-Llama-3.2-1B & 51.31 \\
Claude 4.5 Sonnet & 59.62 & \textbf{68.87} & Skywork-Reward-V2-Qwen3-1.7B & 50.65 \\
doubao-Seed-1.8 & 58.41 & \textbf{70.56} & Skywork-Reward-V2-Llama-3.2-3B & 52.24 \\
DeepSeek-R1 & 70.00 & \textbf{85.79} & Skywork-Reward-V2-Qwen3-4B & 56.26 \\
DeepSeek-V4-Pro & 62.90 & \textbf{71.77} & Skywork-Reward-Llama-3.1-8B-v0.2 & 59.25 \\
DeepSeek-V3 & 59.71 & \textbf{66.82} & Skywork-Reward-V2-Qwen3-8B & 56.16 \\
GLM-5 & 62.61 & \textbf{70.93} & Skywork-Reward-Gemma-2-27B-v0.2 & 61.03 \\
Qwen-Max & 63.64 & \textbf{72.61} & internlm2-1\_8b-reward & 48.87 \\
Qwen3-235B-A22B-Thinking & 63.55 & \textbf{74.95} & Internlm2-7b-reward & 53.55 \\
Qwen3-235B-A22B & 60.09 & \textbf{68.87} &  Internlm2-20b-reward & 57.47 \\
Qwen3-30B-A3B & 57.66 & \textbf{65.51} & ArmoRM-Llama3-8B-v0.1 & 49.43 \\
Qwen3-32B & 58.31 & \textbf{64.20} & Llama-3.1-Nemotron-70B-Reward & 57.94 \\
Qwen3-14B & 50.28 & \textbf{57.94} & DeepSeek-GRM-16B & 61.58 \\
Qwen3-8B & 48.97 & \textbf{57.66} & DeepSeek-GRM-27B & 63.18 \\
Llama-3.3-70B-Instruct & 53.64 & \textbf{59.91} & RM-R1-Qwen2.5-Instruct-7B & 60.74 \\
Llama-3.1-8B-Instruct & 47.01 & \textbf{56.91} & RM-R1-Qwen2.5-Instruct-32B & 62.99 \\
\bottomrule
\end{tabular}%
} 
\vspace{-3pt} 
\captionsetup{justification=raggedright, singlelinecheck=false}
\caption{\textbf{Main Results.} Comparison between model self-evaluation and our \textbf{F$^{2}$DR} preference scores (left), and the performance of other RMs on the \textbf{DeepSearch RM-bench} (right).}
\label{tab:main_result}
\end{table*}

\paragraph{Overall Statistics.}

DeepSearch RM-Bench consists of 1,070 preference pairs built over {1,070 unique queries (one pair per query), covering both English and Chinese. We evaluate the benchmark across three perspectives—task taxonomy, domain distribution, and execution preference—to assess its quality, diversity, and representativeness. We categorize its queries into six task types, ranging from factual retrieval to complex Decision Making. As shown in Figure~\ref{fig:fig_2}, \textit{Domain Knowledge} forms the largest proportion (45.98\%), serving as the primary scenario requiring advanced information synthesis. The benchmark additionally includes General Knowledge, How To guides, Advisory Service, and Subjective Views tasks. It covers 10 vertical domains, with knowledge-intensive sectors (e.g., Science Technology, Finance Economics, Medical Health, Government Law) comprising over 57\%. For query length, we measure Chinese text by characters and English text by words: Chinese queries range from 6 to 286 characters (median 23, mean 31.5), while English queries range from 6 to 85 words (median 12, mean 16.4), and on average one English word corresponds to approximately 1.92 Chinese characters. The overall length distribution is right-skewed—38.0\% of queries are short (6--19 characters), 42.4\% are medium (20--39 characters), and 19.5\% are long (40--286 characters)—with roughly 80\% of queries falling between 6 and 40 characters and a small long tail of complex multi-hop questions extending up to 286 characters. Table~\ref{tab:compare_chosen_rejected} compares execution statistics: chosen trajectories exhibit higher complexity and information density than rejected ones, featuring more planning iterations, frequent retrievals, and longer responses, which indicates that preference signals in DeepSearch are driven by reasoning depth and comprehensive evidence synthesis.

\section{Experiment}
We evaluated 32 models, including advanced closed-source models such as GPT-5,\footnote{\url{https://openai.com/gpt-5/}} Claude 4.5 Sonnet.\footnote{\url{https://www.anthropic.com/claude/sonnet}} For open-source RMs, we mainly evaluated the Skywork series and the RM-R1 series.

\subsection{Main Results}
Our core evaluation metric is \textbf{preference accuracy}, defined as the proportion of samples where the model correctly assigns a higher preference score to the human-annotated \textit{chosen} trajectory than to the \textit{rejected} trajectory. As shown in Table \ref{tab:main_result}, we first compare the performance of our F$^{2}$DR framework against the vanilla self-evaluation method. In the self-evaluation setting, models directly assign a holistic scalar score to the entire DeepSearch workflow, with the specific evaluation prompt detailed in Table \ref{tab:self_evaluation}. This unconstrained scoring paradigm lacks structured dimensional decomposition and process-level auditing, fundamentally hindering the models' ability to grasp the critical evaluation dimensions specific to DeepSearch scenarios.

Under the F$^{2}$DR framework, models successfully overcome these limitations through structured, dimension-wise constraints, achieving a significant leap in effectiveness. On average, F$^{2}$DR yields a 10-percentage-point improvement in preference accuracy, precisely identifying subtle logical inconsistencies and process-level defects that self-evaluation frequently overlooks. The substantial and uniform gains achieved across all frontier models suggest that the primary bottleneck for DeepSearch evaluation lies not in the foundational knowledge capacity of large models, but in the lack of structured guidance and standardized process-level evaluation mechanisms for complex multi-step workflows.

\begin{figure*}[ht]
  \centering
  \includegraphics[width=1\textwidth]{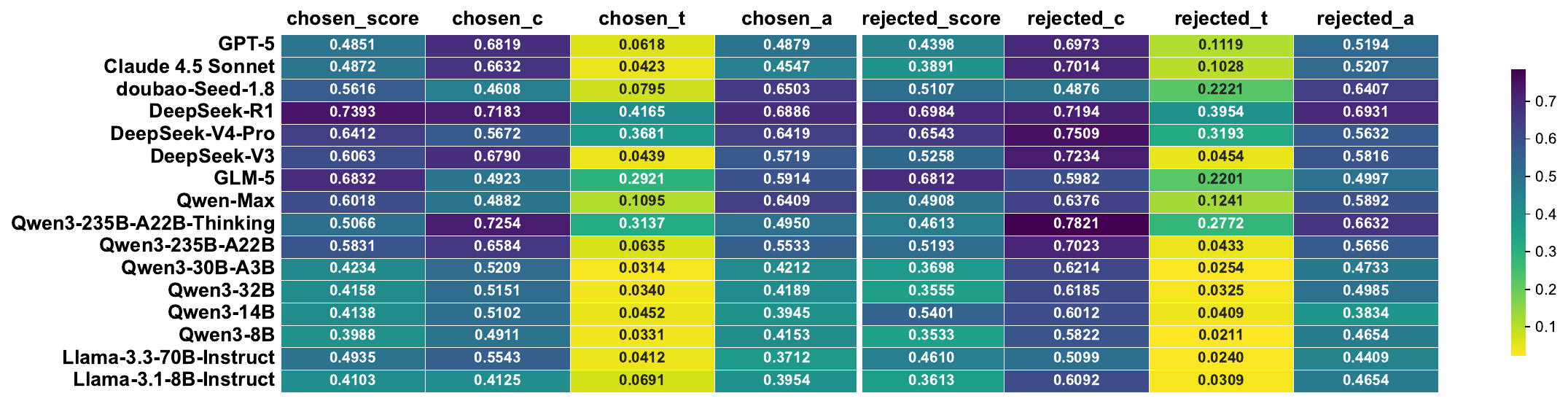}  
  \caption{Pearson Correlation Coefficients of Evaluation Dimensions: Model vs. Human Evaluation under the Unified F$^{2}$DR Framework.}
  \label{fig:fig_3} 
\end{figure*}

Notably, reasoning-centric models exhibit a decisive advantage over traditional foundational models in this complex evaluation task. DeepSeek-R1 not only achieves the highest absolute preference score of 85.79 but also shows the most pronounced improvement of 15.79 points under the F$^{2}$DR framework. This significantly outperforms traditional closed-source models like GPT-5 and Claude 4.5 Sonnet, which achieve scores of 69.25 and 68.87 respectively. Similarly, Qwen3-235B-A22B-Thinking achieves a preference score of 74.95, outperforming its standard counterpart Qwen3-235B-A22B at 68.87 by 6.08 points. These results highlight that F$^{2}$DR effectively unlocks the potential of reasoning models by prompting them to precisely map retrieved information to atomic knowledge at the content level, make metacognitive judgments on planning and reflection for the search trajectory, and execute rigorous consistency checks on the final answer. Furthermore, even the lightweight Qwen3-8B achieves a preference score of 57.66 under F$^{2}$DR, closely approaching the vanilla self-evaluation performance of several top-tier closed-source models. This demonstrates that rigorous, full-pipeline evaluation constraints can effectively compensate for parameter scale limitations to ensure robust evaluation quality.

Comparisons with open-source reward models (RMs) highlight the limitations of traditional RM paradigms. Bradley-Terry discriminative models (e.g., Internlm2-7b-reward: 48.87, ArmoRM-Llama3-8B-v0.1: 49.43) perform poorly, as compressing dynamic, dense DeepSearch trajectories into a single scalar score incurs severe information entropy loss. Although generative reasoning-enhanced models (e.g., RM-R1-Qwen2.5-Instruct-32B: 62.99, DeepSeek-GRM-27B: 63.18) achieve stronger performance via reasoning mechanisms, they evaluate the final answers, neglecting intermediate planning or reflection processes.

\begin{table}[t]
\centering
\fontsize{6.5pt}{7.8pt}\selectfont 
\renewcommand{\arraystretch}{0.85}
\setlength{\tabcolsep}{1.5pt}
\begin{tabular}{l c c c c}
\toprule
\textbf{Models} & \textbf{F$^{2}$DR} & \textbf{w/o Content} & \textbf{w/o Trajectory} & \textbf{w/o Answer} \\
\midrule
GPT-5 & 69.25 & 67.94 & 64.95 & 65.98 \\
Claude 4.5 Sonnet & 68.87 & 62.24 & 67.94 & 61.03 \\
doubao-Seed-1.8 & 70.56 & 67.38 & 69.53 & 65.51 \\
DeepSeek-R1 & 85.79 & 80.28 & 78.41 & 81.30 \\
DeepSeek-V4-Pro & 71.77 & 70.93 & 69.34 & 67.01 \\
DeepSeek-V3 & 66.82 & 60.00 & 64.57 & 62.42 \\
GLM-5 & 70.93 & 67.19 & 69.87 & 70.37 \\
Qwen-Max & 72.61 & 69.07 & 67.38 & 70.84 \\
Qwen3-235B-A22B-Thinking & 74.95 & 73.45 & 70.18 & 70.84 \\
Qwen3-235B-A22B & 68.87 & 64.57 & 66.07 & 63.17 \\
Qwen3-30B-A3B & 65.51 & 62.17 & 64.58 & 64.85 \\
Qwen3-32B & 64.20 & 63.92 & 61.02 & 58.97 \\
Qwen3-14B & 57.94 & 57.01 & 56.36 & 54.20 \\
Qwen3-8B & 57.66 & 54.95 & 57.19 & 54.95 \\
Llama-3.3-70B-Instruct & 59.91 & 55.42 & 56.16 & 56.82 \\
Llama-3.1-8B-Instruct & 56.91 & 53.45 & 55.14 & 55.38 \\
\bottomrule
\end{tabular}
\caption{Ablation results of F$^{2}$DR framework across different models.}
\label{tab:ablation_study}
\end{table}

\begin{figure*}[ht]
  \centering
  \includegraphics[width=1\textwidth]{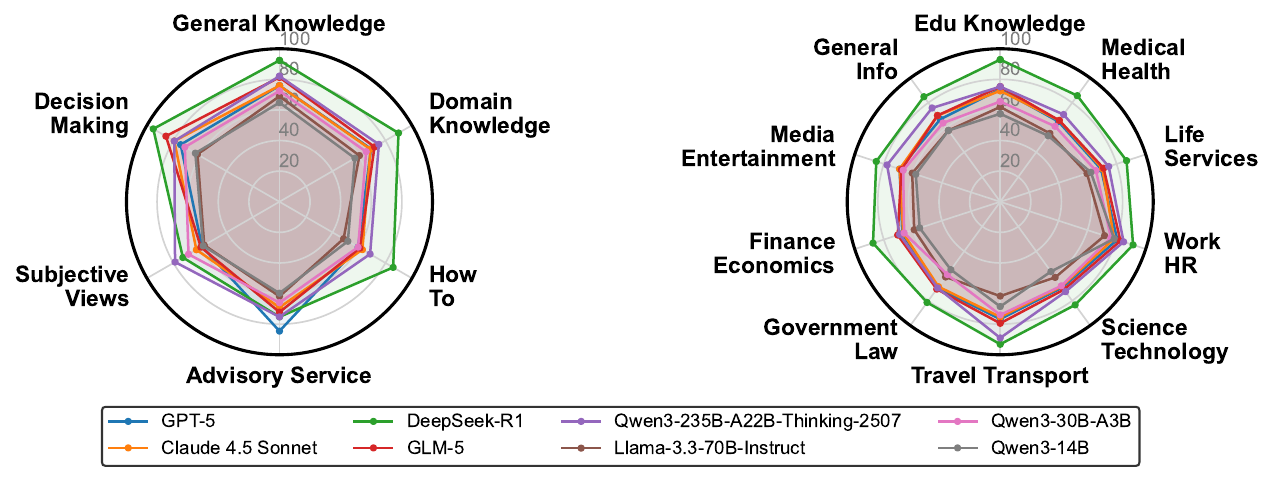}  
  \caption{Performance of various models across task and domain dimensions on the DeepSearch RM-bench.}
  \label{fig:fig_4} 
\end{figure*} 

\subsection{Ablation Study}

As shown in Table \ref{tab:ablation_study}, ablation studies validate the independent contributions of the F$^{2}$DR framework's three dimensions: removing any single dimension systematically degrades performance across all models, confirming their complementarity. Model architectures exhibit distinct sensitivities to these dimensions. Thinking models with explicit reasoning are most affected by the trajectory dimension (missing this dimension causes DeepSeek-R1 to drop by 7.38 points and Qwen3-235B-A22B-Thinking by 4.77 points), underscoring the necessity of process-level evaluation. Conversely, non-Thinking models rely heavily on the answer dimension (Claude 4.5 Sonnet drops by 7.84 points), revealing their inability to effectively evaluate the trajectory dimension. Furthermore, the content relevance dimension provides a stable, foundational contribution, as its removal consistently leads to varying degrees of performance degradation across all models. Overall, F$^{2}$DR bridges outcome quality and reasoning process evaluation, establishing a fairer, more comprehensive standard for assessing DeepSearch capabilities across diverse model architectures.

\subsection{Analysis of Evaluation Dimensions}

We further measured the consistency between model judgments and annotated evaluation dimensions using the Pearson correlation coefficient. Detailed results are shown in Figure \ref{fig:fig_3}. Notably, Thinking-series models with enhanced reasoning capabilities generally exhibit more prominent consistency across all dimensions, demonstrating stronger alignment with the F$^{2}$DR criteria. From a dimension perspective, the Content and Answer dimensions exhibited relatively high consistency scores. This is mainly due to their intuitive evaluation logic and reliance on quantifiable checklist-based objective verification, enabling models to reliably capture the core characteristics of evaluation criteria. In contrast, the Trajectory dimension generally exhibits lower consistency. This limitation mainly stems from the models' insufficient understanding of subjective evaluation signals, as well as the difficulty of accurately identifying and localizing defects within the planning-reflection process.

\begin{table}[!t]
\centering
\small  
\resizebox{\linewidth}{!}{
\begin{tabular}{l c c}
\toprule
\textbf{Model} & \textbf{Chosen Avg} & \textbf{Rejected Avg} \\
\midrule
GPT-5                         & 0.5902 & 0.5034 \\
Claude 4.5 Sonnet             & 0.2810 & 0.2363 \\
doubao-Seed-1.8               & 0.9733 & 0.9374 \\
DeepSeek-R1                   & 0.8934 & 0.8123 \\
DeepSeek-V4-Pro               & 0.9334 & 0.8931 \\
DeepSeek-V3                   & 0.9683 & 0.9607 \\
GLM-5                         & 0.9504 & 0.9003 \\
Qwen-Max                      & 0.9351 & 0.8455 \\
Qwen3-235B-A22B-Thinking & 0.8371 & 0.6928 \\
Qwen3-235B-A22B               & 0.9344 & 0.9029 \\
Qwen3-30B-A3B                 & 0.9561 & 0.9415 \\
Qwen3-32B                     & 0.9309 & 0.9012 \\
Qwen3-14B                     & 0.9543 & 0.9398 \\
Qwen3-8B                      & 0.9845 & 0.9552 \\
Llama-3.3-70B-Instruct        & 0.8898 & 0.7523 \\
Llama-3.1-8B-Instruct         & 0.8423 & 0.7752 \\
\bottomrule
\end{tabular}
}
\caption{Average Scores on Chosen and Rejected Trajectory Dimensions.}
\label{tab:trajectory_dimensions}
\end{table}

\subsection{Analysis of Trajectory Evaluation}

In our fine-grained Trajectory dimension analysis, we evaluated model scores against the human-annotated ground truth. As shown in Table \ref{tab:trajectory_dimensions}, Thinking-series models demonstrate superior discrimination. Notably, Qwen3-235B-A22B-Thinking and DeepSeek-R1 stand out with score gaps of 0.1443 and 0.0811, respectively, effectively identifying specific flaws like keyword redundancy and unstructured reasoning within the planning and reflection loop. Several non-Thinking-series models also show competitive evaluation capabilities, including Qwen-Max (0.0896 gap), GLM-5 (0.0501), and DeepSeek-V4-Pro (0.0403). Conversely, DeepSeek-V3, Qwen3-8B, and Doubao-Seed-1.8 consistently score near 1.0 across all trajectories; this negligible gap reveals severe calibration issues and an inability to differentiate planning quality. Furthermore, GPT-5 and Claude 4.5 Sonnet score very low, also with marginal gaps, suggesting a general inability among most large closed-source models to reliably evaluate complex DeepSearch trajectories.

\subsection{Analysis of Query Type}
We systematically evaluated mainstream LLMs across six general dimensions and ten industrial domains, revealing a distinct performance stratification. The results are illustrated in Figure \ref{fig:fig_4}. DeepSeek-R1 maintains a lead across all evaluated metrics, attributed to its superior reasoning capabilities. In contrast, the performance of other models aligns with the scaling law, where evaluative proficiency improves with increasing parameter scale.
The domain-level analysis reveals that models exhibit high evaluative consistency in daily scenarios such as Work HR and Travel Transport. This proficiency stems from the high data density and standardized logic in pre-training corpora, allowing for precise pattern matching. However, significant performance degradation occurs in high-threshold domains like Medical Health and Science Technology, indicating that general-purpose capabilities remain insufficient to bridge the gap toward specialist-level cognition. The How To (or procedural knowledge) dimension represents a universal evaluative bottleneck. Models are proficient in declarative descriptions focused on what to say but struggle with procedural logic pertaining to how to act. Lacking a grounded understanding of causality and temporal sequences, they tend to judge quality based on linguistic probability rather than identifying latent logical breaks in execution steps.

\section{Conclusion}
To address the critical absence of process-level evaluation for DeepSearch scenarios, we present F$^{2}$DR and DeepSearch RM-Bench. F$^{2}$DR assesses the entire DeepSearch workflow across three complementary fine-grained dimensions: Content, Trajectory, and Answer. Built upon this framework, DeepSearch RM-Bench is the first dedicated benchmark for evaluating reward models in full-pipeline DeepSearch scenarios, comprising high-quality preference pairs annotated with rigorous process-level labels. Experimental results show that F$^{2}$DR significantly outperforms existing baselines in evaluation consistency. Furthermore, DeepSearch RM-Bench poses significant challenges to mainstream open-source reward models, revealing their fundamental performance bottlenecks in complex multi-step decision-making chains.

\section*{Acknowledgments}
The present research was supported by the National Key Research and Development Program of China (Grant No. 2024YFE0203000), the China Postdoctoral Science Foundation (Grant No. 2026M791664) and the Postdoctoral Fellowship Program of CPSF (Grant No. GZC20251075). We would like to thank the anonymous reviewers for their insightful comments.

\section*{Limitation}
While current evaluations depend on API calls, optimizing the performance-efficiency trade-off is our next step. By training a dedicated DeepSearch RM using fine-grained F$^{2}$DR signals as supervision, we can substantially lower resource costs without compromising complex search logic modeling.

\bibliography{custom}
\clearpage
\onecolumn
\appendix

\section{Case Study for DeepSearch Workflow}
\label{appendixA}
We present a user query and the corresponding response generated under the DeepSearch framework, which allows us to clearly understand how DeepSearch addresses complex user queries. Due to space constraints, we provide a concise summary of the planning and reflection, information retrieval, and final summarization stages.

\begin{table}[htbp]
  \centering
  \small
  \begin{tabular}{p{0.18\linewidth} p{0.78\linewidth}}
    \toprule
    \textbf{Component} & \textbf{Content} \\
    \midrule
    \textbf{User Query} & How can I learn the language and pass the IELTS exam to work as a care worker in Australia? Please help me outline a detailed plan, investment costs, and important considerations. \\
    \midrule
    \textbf{Round 1} & 
    \textit{Thinking \& Planning}: Identify the differences between care worker and registered nurse requirements, and check language thresholds. Plan to search for care worker English requirements, study plans, and initial visa matters. \newline
    \textit{Recall Information}: 
    1. Clear threshold: Care workers (482 visa) only need an IELTS score of 5.0 (no subscore below 4.5 or 5.0), much lower than the 7.0 required for registered nurses. 
    2. Basic study framework: A 3-month preparation plan covering vocabulary and section-specific practice. \newline
    \textit{Summary}: Established a study goal centered on an IELTS score of 5.0 and built initial preparation logic. \\
    \midrule
    \textbf{Round 2} & 
    \textit{Thinking \& Planning}: Deepen exploration of cost gaps (Task 3 results were empty) and career entry details. Plan to search for 2026 updated exam fees, training costs, and Certificate III course details. \newline
    \textit{Recall Information}: 
    1. 2026 cost update: IELTS exam fee reduced to 1990 RMB; PTE approximately 2255 RMB. 
    2. Clear qualification: Entry requires a Certificate III in Individual Support, with a study duration of 6–9 months. 
    3. Living costs: Rent and expenses are approximately 300–500 AUD per week. \newline
    \textit{Summary}: Filled in key facts on financial budgets and industry entry qualifications (Cert III). \\
    \midrule
    \textbf{Round 3} & 
    \textit{Thinking \& Planning}: Correct the assessment authority (exclude ANMAC) and complete the visa application loop. Plan to search for ACWA assessment processes, compile visa checklists, and practical considerations. \newline
    \textit{Recall Information}: 
    1. Authority correction: Care worker (ANZSCO 423111) skill assessment is primarily handled by ACWA. 
    2. Visa checklist: 482 visa requires employer sponsorship, skill assessment, medical examination, and police clearance. 
    3. Work reality: Annual salary approximately 330,000 RMB; full-time work requires at least 5 days per week. \newline
    \textit{Summary}: Constructed a full-chain execution map from language preparation to skill assessment, visa application, and life in Australia. \\
    \midrule
    \textbf{Final Answer} & 
    1. \textbf{Language threshold}: IELTS overall and subscores must reach 5.0 (or equivalent PTE score), which is a requirement for visa and employment. \newline
    2. \textbf{Core qualification}: Must obtain an Australian Certificate III in Individual Support, including at least 120 hours of mandatory practical placement. \newline
    3. \textbf{Visa pathway}: Primarily enter via the 482 Employer Sponsored Visa, with three steps: employer sponsorship, position nomination, and individual visa application. \newline
    4. \textbf{Cost-benefit}: Self-study costs start at approximately 3,000 RMB; annual salary is about 50,000 AUD ($\approx$230,000 RMB). Be prepared for high-intensity physical work. \\
    \bottomrule
  \end{tabular}
  \caption{Case Study: DeepSearch Workflow for Australian Care Worker Query.}
  \label{tab:case_study_care_worker}
\end{table}

\clearpage

\section{Prompts for F$^{2}$DR}
\label{appendixB}
\subsection{Prompts for Content dimension}
This prompt is the Content-dimension evaluation template in F$^{2}$DR, using checklist-based scoring  and JSON output to ensure consistent, objective assessments of DeepSearch content dimension.
\begin{table}[h]
    \centering
    \small 
    \begin{tabular}{p{0.95\columnwidth}}
    \hline
    \\
    \textbf{\#\# Persona} \\
    You are a \textbf{Senior Information Coverage Assessment Expert}, specializing in evaluating the accuracy of Retrieved Content (Reference) against key Information Points (Checklist) derived from the User Query. \\
    \\
    \textbf{\#\# Objective} \\
    Based on the provided User Query, Retrieved References, and the Preset Information Checklist, analyze the coverage degree of each information point individually and provide an authoritative score. \\
    \\
    \textbf{\#\# Scoring Criteria} \\
    You must strictly adhere to the following standards: \\
    \quad - \textbf{1.5 points (Fully Covered)}: The retrieved material contains the core facts of the information point, and the data/content in brackets is also consistent. \\
    \quad - \textbf{1.0 point (Basically Covered)}: The retrieved material contains the core facts of the information point. \\
    \quad - \textbf{0.5 points (Partially Covered)}: The retrieved material covers part of the content or mentions relevant information. \\
    \quad - \textbf{0 points (Not Covered)}: The retrieved material completely fails to mention this point, or the content logically conflicts with it. \\
    \quad - \textbf{Note}: Assign 0 points only if \textit{no} references mention the current information point. \\
    \\
    \textbf{\#\# Task Requirements} \\
    1. \textbf{Deep Discrimination}: Ignore surface keyword matching and focus on content semantics. Exclude noise in search results that does not align with the user's original intent. \\
    2. \textbf{Classification}: \\
    \quad - \textbf{Main (Primary Points)}: The ``cornerstone'' of the answer; must be rigorously assessed. \\
    \quad - \textbf{Second (Secondary Points)}: Supplementary background; assess its contribution to professional depth. \\
    3. \textbf{Reasoning Process}: Before outputting JSON, first summarize the content of each ref, then reason one by one whether each information point is covered. \\
    4. \textbf{Output Restriction}: Finally, output a standard JSON array. \textbf{Strictly prohibit} including any leading words or explanations. \\
    \\
    \textbf{\#\# Input Information} \\
    \textbf{User Query}: \{Query\} \\
    \textbf{Required Checklist}: \{Check\_list\} \\
    \textbf{Retrieved Refs}: \{All\_refs\} \\
    \\
    \textbf{\#\# Output Format} \\
    \texttt{<crm>} \\
    Summarize content of each ref: xxx \\
    Main Point 1: 1.5, Reason: xxx \\
    Main Point 2: 0.5, Reason: xxx \\
    ... \\
    Second Point 1: 0, Reason: xxx \\
    \texttt{</crm>} \\
    \texttt{<score>} \\
    \{\{``main1'': 1.5, ``main2'': 0.5, ..., ``second1'': 0, ...\}\} \\
    \texttt{</score>} \\
    \\
    Now, please begin your assessment. \\
    \hline
    \end{tabular}
    \caption{Prompt for the Content Dimension in F$^{2}$DR.}
    \label{tab:coverage_prompt}
\end{table}

\clearpage

\subsection{Prompts for Trajectory dimension}
This prompt is the Trajectory-dimension evaluation template in F$^{2}$DR, using multi-dimensional binary scoring to assess AI planning trajectory rationality and output structured results.
\begin{table}[h]
    \centering
    \small
    \begin{tabular}{p{0.95\textwidth}}
    \hline
    \\
    \textbf{\#\# Persona} \\
    You are a \textbf{Rigorously Strict Search Strategy Evaluation Expert}. You must evaluate the planning reasoning process and the rationality of the query planning trajectory generated by the AI assistant with \textbf{extreme objectivity and rigor}. 
    \textbf{Evaluation Objects} \\
    1. \textbf{Reasoning Process}: The complete reasoning on ``how to answer'' and ``how to plan'' after receiving the user query. \\
    2. \textbf{Planning Trajectory}: A DAG list of sub-tasks. Each sub-task is a JSON object (focus only on \texttt{search}/\texttt{mcp} types; \\
    \quad - \texttt{id}: Unique identifier (starts from '1'). \\
    \quad - \texttt{type}: 'search', 'code', 'mcp'. \\
    \quad - \texttt{para}: Keywords list (search/mcp) or type description (code). \\
    \quad - \texttt{desc}: Clear guidance for execution. \\
    \textbf{Evaluation Dimensions \& Rules} \\
    \textbf{Scale}: Only \textbf{1 (Pass)} or \textbf{0 (Fail)}. Any deviation results in 0. Focus heavily on \textbf{Notes} (common errors).\\
    \textbf{I. Task Level Dimensions} \\
    \textbf{1. Intent Understanding Correctness} \\
    - \textbf{Criteria}: Exact match between reasoning and user intent. Must identify explicit needs and all implicit constraints (timeliness, professional depth, data range). \\
    - \textbf{Note}: Check if hidden user needs are satisfied. \\
    \textbf{2. Search Task Atomicity} \\
    - \textbf{Criteria}: Each keyword in \texttt{para} must be a semantic unit. No multi-entity/cross-domain searches in one keyword. \\
    - \textbf{Exception}: Using the original user query verbatim is allowed. \\
    \textbf{3. Reasoning-Planning Consistency} \\
    - \textbf{Criteria}: Every assumption/gap in reasoning must map to a specific task in the trajectory. \\
    - \textbf{Note}: Watch for cases where reasoning identifies gaps but the DAG is empty. \\
    \textbf{II. Trajectory Level Dimensions} \\
    \textbf{4. Gap Resolution Degree} \\
    - \textbf{Criteria}: Clear trend of \textbf{entropy reduction}. Each task solves a specific gap. No redundancy. \\
    - \textbf{Note}: Repetitive reasoning is NOT entropy reduction. \\
    \textbf{5. Planning Comprehensiveness} \\
    - \textbf{Criteria}: Covers all explicit and implicit information dimensions (core and secondary) without omission. \\
    - \textbf{Note}: Watch for missing information dimensions. \\
    \textbf{6. Reasoning Iteration Innovation} \\
    - \textbf{Criteria}: In multi-round planning, each round must offer incremental value/optimization. No exact repetition. Secondary searches must require new keywords/angles. \\
    - \textbf{Note}: Watch for identical reasoning content across rounds. \\
    \textbf{\#\# Input Information} \\
    \textbf{User Query}: \{query\} \\
    \textbf{Planning History}: \{plan\_search\_history\} \\
    \textbf{\#\# Output Format} \\
    Encapsulate results in tags. First summarize the reasoning, then score each dimension. \\
    \texttt{<trm>} \\
    Reasoning Summary: ... \\
    Intent Understanding: [1 or 0], Reason: ... \\
    Search Atomicity: [1 or 0], Reason: ... \\
    Consistency: [1 or 0], Reason: ... \\
    Gap Resolution: [1 or 0], Reason: ... \\
    Comprehensiveness: [1 or 0], Reason: ... \\
    Iteration Innovation: [1 or 0], Reason: ... \\
    \texttt{</trm>} \\
    \texttt{<score>} {[1, 1, 0, 1, 0, 1]} \% Python List Format {</score>} \\
    Now, please begin your assessment. \\
    \hline
    \end{tabular}
    \caption{The prompt for evaluating the Trajectory Dimension in F$^{2}$DR.}
    \label{tab:trajectory_prompt}
\end{table}

\clearpage

\subsection{Prompts for Answer dimension}
This prompt is the Answer-dimension evaluation template in F$^{2}$DR, using three-tier scoring and structured output to assess the quality of final answers in DeepSearch.

\begin{table}[h]
    \centering
    \small
    \begin{tabular}{p{0.95\textwidth}}
    \hline
    \\
    \textbf{\#\# Persona} \\
    You are a \textbf{Rigorously Strict Search Strategy Evaluation Expert}. You must evaluate the final answer generated by the query planning AI assistant based on the user query with \textbf{extreme objectivity and rigor}. Apply high standards throughout the scoring; do not award 1 point unless absolutely necessary. \\
    \textbf{General Evaluation Rules} \\
    1. \textbf{Three-Tier Scoring}: All dimensions (subjective and objective) use three tiers: \textbf{1 (Excellent)}, \textbf{0.5 (Acceptable)}, \textbf{0 (Unqualified)}. \\
    2. \textbf{Independent Evaluation}: When assessing subjective dimensions (Dimensions 1–3), do NOT refer to checklist items; judge only against the original user query. \\
    3. \textbf{Precise Localization}: When scoring 0.5 or 0, clearly locate the exact problem in the reasoning. \\
    \textbf{I. Subjective Evaluation Dimensions (1 / 0.5 / 0)} \\
    \textbf{Dimension 1: Demand Understanding} \\
    - Core Criterion: Matching degree and full coverage of the answer to user demand. Do NOT refer to checklist items for this dimension. \\
    - 1 (Excellent): Fully understands core user demand, covers all explicit key points, accurately identifies and satisfies potential implicit intent, with no omissions. \\
    - 0.5 (Acceptable): Basically understands demand, fully covers explicit key points, but insufficiently identifies potential implicit intent (e.g., background supplementation, comparison needs) or has minor understanding deviations. \\
    - 0 (Unqualified): Misses core explicit demand or contains severe intent recognition errors. \\
    \textbf{Dimension 2: Content Quality} \\
    - Core Criterion: Completeness, conciseness, and logicality of the answer. Do NOT refer to checklist items for this dimension. \\
    - 1 (Excellent): Detailed and substantial; extremely concise without redundancy; clearly focused with high information gain; perfectly logical and consistent. \\
    - 0.5 (Acceptable): Basically complete but has minor flaws such as slight redundancy, minor repetition, unclear prioritization, or slightly rigid logical derivation. \\
    - 0 (Unqualified): Contains contradictions, massive repetition, chaotic logic, or severely insufficient information depth. \\
    \textbf{II. Objective Evaluation Dimensions (1 / 0.5 / 0)} \\
    \textbf{Dimension 3: Answer Coverage Rate} \\
    - Evaluation Basis: Based on the provided checklist, evaluate coverage of \textbf{core information (Main)} only. Check only whether entity information is hit; ignore content in parentheses. \\
    - Three-Tier Criteria: \\
    \quad 1: Core information (Main) hit rate $\geq 60\%$ \\
    \quad 0.5: $25\% \leq \text{Core information (Main) hit rate} < 60\%$ \\
    \quad 0: Core information (Main) hit rate $< 25\%$ \\
    \textbf{\#\# Input Information} \\
    \textbf{User Query}: \{query\} \\
    \textbf{Required Information List}: \{checklist\} \\
    \textbf{Final Model Answer}: \{answer\} \\
    \textbf{\#\# Output Format Requirements} \\
    \texttt{<arm>} \\
    Answer Summary: xxx \\
    Demand Understanding: [1/0.5/0], Reason: [Evaluation based on Demand Understanding criteria] \\
    Content Quality: [1/0.5/0], Reason: [Evaluation based on Content Quality criteria] \\
    Answer Coverage: [1/0.5/0], Reason: [Specific hit ratio of core/supplementary info against checklist] \\
    \texttt{</arm>} \\
    \texttt{<score>} {[1, 0.5, 0.5]} \% Python List Format {</score>} \\
    Now, please begin your assessment. \\
    \hline
    \end{tabular}
    \caption{The prompt for evaluating the Answer Result Dimension in F$^{2}$DR.}
    \label{tab:answer_prompt}
\end{table}

\clearpage
\twocolumn 

\section{Prompts for Self-evaluation method} 
\label{appendixC}
This is the prompt for the self-evaluation prompt.

\begin{table}[h]
    \centering
    \small
    \begin{tabular}{p{0.95\linewidth}} 
    \hline
    \\
    \textbf{\#\# Persona} \\
    You are a \textbf{Rigorously Strict Search Strategy Evaluation Expert}. You must evaluate the planning reasoning process, the rationality of the planning trajectory, and the quality of the final answer generated by the query planning AI assistant based on the user query, with \textbf{extreme objectivity and rigor}. You need to formulate your own scoring criteria to assign a score. \\
    \textbf{Evaluation Task} \\
    Evaluate the search effectiveness and final answer quality by judging the rationality of the search plan and adequacy of retrieved information, verifying whether the final answer can accurately respond to the user query, and assigning a decimal score between 0 and 1. \\
    \textbf{Input Information} \\
    \textbf{User Query}: \{query\} \\
    \textbf{Search Plan \& History}: \{history\_str\} \\
    \textbf{Final Answer}: \{answer\} \\
    \textbf{Output Format} \\
    Please strictly follow the format below: \\
    Detailed evaluation reason \\
    \texttt{<score>}Score\texttt{</score>} \\
    \hline
    \end{tabular}
    \caption{The prompt for self-evaluation.}
    \label{tab:comprehensive_eval_prompt}
\end{table}

\section{Case Study for F$^{2}$DR}
\label{appendixD}
There is a case study on the checklist for user queries, the three evaluation dimensions of F$^{2}$DR.
\begin{table}[htbp]
  \centering
  \small
  \begin{tabular}{p{0.20\linewidth} p{0.76\linewidth}}
    \toprule
    \textbf{Component} & \textbf{Content} \\
    \midrule
    \textbf{User Query} & How can I learn the language and pass the IELTS exam to work as a care worker in Australia? Please help me outline a detailed plan, investment costs, and important considerations. \\
    \midrule
    \textbf{Main Checklist Items} & 
    1. IELTS General Training (General) test type \newline
    2. Basic IELTS requirement for Australian care workers (overall 5.0, no subscore below 5.0) \newline
    \dots (omitted) \\ 
    \midrule
    \textbf{Secondary Checklist Items} & 
    1. IELTS requirement for Chinese employers \newline
    2. Re-examination fee (2,170 RMB) \newline
    \dots (omitted) \\
    \midrule
    \textbf{Content Evaluation} & 
    \textit{Reference Summary}: \newline
    - refs\_ID 1--3: Cover IELTS requirement adjustments for Australian nursing occupations \dots (omitted)\newline
    \textit{Main Point 1 (IELTS General Training test type)}: 0.5 points. Reason: \newline
    \dots (omitted) \\ 
    \bottomrule
  \end{tabular}
  \caption{Checklist and Content Dimensional Evaluation for Australian Care Worker Query.}
  \label{tab:content}
\end{table}

\begin{table}[htbp]
  \centering
  \small
  \begin{tabular}{p{0.18\linewidth} p{0.78\linewidth}}
    \toprule
    \textbf{Component} & \textbf{Content} \\
    \midrule
    \textbf{User Query} & How can I learn the language and pass the IELTS exam to work as a care worker in Australia? Please help me outline a detailed plan, investment costs, and important considerations. \\
    \midrule
    \textbf{Trajectory Evaluation} & 
    \textit{Evaluation Results} \newline
    \textit{Summary of Thinking Process}: \newline
    In response to the user's query ``How can I learn the language and pass the IELTS exam to work as a care worker in Australia? \dots (omitted)\newline
    - \textbf{Round 1}: Identify core needs (English requirements for care workers, IELTS study plan, cost estimation, precautions), and plan 4 independent search tasks covering occupational requirements, preparation resources, fees, and visa information. \newline
    \dots (omitted)\newline
    \textit{Intent Understanding Correctness}: [1 point]. Reason: Accurately identify explicit demands (IELTS study plan,\dots (omitted)\newline
    \textit{Search Task Atomicity}: [1 point]. Reason: All keywords in search task paragraphs are semantically complete and indivisible \dots (omitted)\newline
    \textit{Reasoning-Planning Consistency}: [1 point]. Reason: Key assumptions in thinking (e.g., ``costs need to estimate course fees and exam fees'') and data gaps (e.g., ``unclear skills assessment authority'')\dots (omitted)\newline
    \textit{Gap Resolution Degree}: [1 point]. Reason: The planning process shows a significant entropy reduction trend—the first round locates basic needs,\dots (omitted)\newline
    \textit{Reasoning Iteration Innovation}: [1 point]. Reason: It fully covers explicit dimensions (language study plan, cost, precautions)\dots (omitted)\newline
    \textit{Innovation of Thinking Iteration}: [1 point]. Reason: Each round of thinking achieves substantial optimization—the second round adjusts keywords based on the first round results\dots (omitted) \\
    \bottomrule
  \end{tabular}
  \caption{User Query and Trajectory Dimensional Evaluation for Australian Care Worker Query.}
  \label{tab:trajectory}
\end{table}

\begin{table}[htbp]
  \centering
  \small
  \begin{tabular}{p{0.18\linewidth} p{0.78\linewidth}}
    \toprule
    \textbf{Component} & \textbf{Content} \\
    \midrule
    \textbf{User Query} & How can I learn the language and pass the IELTS exam to work as a care worker in Australia? Please help me outline a detailed plan, investment costs, and important considerations. \\
    \midrule
    \textbf{Answer Evaluation} & 
    <arm> \newline
    \textit{Summary of Answer Content}: The model's answer provides a comprehensive guide to language learning and IELTS preparation for working as a care worker in Australia,\dots (omitted)\newline
    \textit{Demand Understanding}: [1 point]. Reason:\dots (omitted)\newline
    \textit{Content Quality}: [1 point]. Reason: The content is detailed and complete (phased plan table, \dots (omitted)\newline
    \textit{Answer Coverage}: [0.5 points]. Reason: The hit rate of core information (Main) is 50\% (7/14). \dots (omitted)</arm> \newline
    <score> [1, 1, 0.5]</score> \\
    \bottomrule
  \end{tabular}
  \caption{User Query and Answer Dimensional Evaluation for Australian Care Worker Query.}
  \label{tab:answer}
\end{table}

\clearpage

\begin{table}[htbp]
  \centering
  \small
  \begin{tabular}{>{\raggedright\arraybackslash}p{0.18\linewidth} >{\raggedright\arraybackslash}p{0.78\linewidth}}
    \toprule
    \textbf{Component} & \textbf{Content} \\
    \midrule
    \textbf{User Query} & How can I learn the language and pass the IELTS exam to work as a care worker in Australia? Please help me outline a detailed plan, investment costs, and important considerations. \\
    \midrule
    \textbf{Self-evaluation} & 
    1. \textit{Rationality of Search Planning}: The model adopted a 4-round progressive search strategy,\dots (omitted)\newline
    2. \textit{Completeness of Information Acquisition}: Successfully obtained core information such as care worker-specific IELTS score requirements \dots (omitted)\newline
    3. \textit{Quality of Final Answer}: The answer has a clear structure, framed as \dots (omitted)\newline
    \newline
    <score>0.93</score> \\
    \bottomrule
  \end{tabular}
  \caption{User Query and Self-evaluation result for Australian Care Worker Query.}
  \label{tab:self_evaluation}
\end{table}

\section{GSB evaluation}
\label{appendixE}
The industrial-grade human evaluation system adopted for GSB validation consists of 5 rating levels, centering on five core dimensions: intent understanding, content accuracy, information value, professionalism, and scenario adaptability. The detailed grading criteria are as follows:

\begin{itemize}
    \item \textbf{4 (Excellent)}: Accurately understands all user intents and perfectly addresses the problem; content is accurate, timely, and authoritatively traceable, rich in actionable information; well-written with no missing information, high professionalism supported by case data, clear decision-making, appropriate format, and natural conversational tone.
    
    \item \textbf{3 (Satisfactory)}: Accurately understands core intents and meets primary requirements; content is generally accurate, meets timeliness and authority standards, and contains no invalid information; coherent expression, complete information, basic professionalism, clear decision-making, and smooth conversation.
    
    \item \textbf{2 (Marginally Satisfactory)}: Basically understands intents and only meets basic requirements; main content is correct but lacks timeliness and authority, with minor deviations and redundancy; no critical information missing, average professionalism, no case support, and thin content.
    
    \item \textbf{1 (Poor)}: Only marginally aligns with intents and fails to effectively solve the problem; content contains numerous errors, lacks timeliness and authority, and provides little useful information; critical information missing, no professionalism, inappropriate format, poor conversational quality, and memory errors.
    
    \item \textbf{0 (Unsatisfactory or Red Line)}: Completely deviates from intents and provides irrelevant responses; content is false or erroneous, violates legal red lines, outputs negative or inappropriate information, and has no valid value.
\end{itemize}

\begin{table}[htbp] 
\centering
\vspace{0.5em}
\small 
\begin{tabular}{lccc} 
\toprule
\textbf{Annotators} & \textbf{Good} & \textbf{Simple} & \textbf{Bad} \\
\midrule
Annotator 1 & 212 & 5 & 1 \\
Annotator 2 & 210 & 2 & 3 \\
Annotator 3 & 224 & 5 & 3 \\
Annotator 4 & 209 & 3 & 2 \\
Annotator 5 & 215 & 6 & 4 \\
\bottomrule
\end{tabular}
\caption{Inter-Annotator Rating Distribution}
\label{tab:annotator_consistency}
\end{table}

The rating results from the annotators, scored in accordance with the specified criteria, are presented in Table \ref{tab:annotator_consistency}.

\section{Supplemental Information for Experiment}

\label{appendixF}

\subsection{Study on F$^{2}$DR Evaluation Dimension Weights}

We conduct a F$^{2}$DR dimension weight ablation study (Table 16) on DeepSeek-R1. Evaluating all {0.75,1.00,1.25} permutations across three dimensions, our setting (0.75:1.00:1.25) achieves the highest 85.79\% match rate, validating the design.

\begin{table}[htbp]
\centering
\vspace{0.5em}
\scriptsize 
\setlength{\tabcolsep}{2pt} 
\begin{tabular}{lccccc}
\toprule
\textbf{Weights} & \textbf{\makecell{DeepSeek\\-R1}} & \textbf{\makecell{Qwen3-235B\\-A22B\\-Thinking}} & \textbf{GPT-5} & \textbf{\makecell{Claude 4.5\\Sonnet}} & \textbf{\makecell{doubao\\-Seed-1.8}} \\
\midrule
setting1 1.00:1.00:1.00 & 83.52 & 73.75 & 68.93 & 68.47 & \textbf{71.03} \\
setting2 0.75:1.25:1.00 & 84.76 & 74.87 & 68.23 & 67.82 & 70.61 \\
setting3 1.00:0.75:1.25 & 81.96 & 73.21 & 67.45 & 66.16 & 65.98 \\
setting4 1.00:1.25:0.75 & 84.57 & 72.94 & 67.30 & 64.09 & 68.24 \\
setting5 1.25:0.75:1.00 & 83.83 & 74.05 & 64.17 & 65.28 & 67.98 \\
setting6 1.25:1.00:0.75 & 84.30 & 72.11 & 65.34 & 66.67 & 70.46 \\
\textbf{setting7 0.75:1.00:1.25} & \textbf{85.79} & \textbf{74.95} & \textbf{69.25} & \textbf{68.87} & 70.56 \\
\bottomrule
\end{tabular}
\caption{Study on F$^{2}$DR Dimension Weights Across Models.}
\label{tab:dimension_weight}
\end{table}

\section{Annotator Information}

\label{appendixG}
Our annotators are all in-house employees. We have a total of 5 annotators, each earning 500 RMB per day. On average, each staff member produces around 100 data entries daily, and the entire annotation period lasts approximately 10 days. They are mainly responsible for preference-based data verification, as well as scoring and filtering of GSB-based data.

\end{document}